\documentclass{article}

\usepackage[final]{nips_2018}

\usepackage{natbib}
\usepackage{graphicx}
\usepackage[utf8]{inputenc} 
\usepackage[T1]{fontenc}    
\usepackage{hyperref}       
\usepackage{url}            
\usepackage{booktabs}       
\usepackage{amsfonts}       
\usepackage{nicefrac}       
\usepackage{microtype}      

\title{Explain to Fix: A Framework to Interpret and Correct {DNN} Object Detector Predictions}

\author{
  Denis~Gudovskiy, Alec~Hodgkinson\\
  Panasonic Beta Research Lab, Mountain View, CA, 94043, USA \\
  \texttt{\{denis.gudovskiy, alec.hodgkinson\}@us.panasonic.com} \\
  \And
  Takuya~Yamaguchi, Yasunori~Ishii, Sotaro~Tsukizawa \\
  Panasonic AI Solutions Center, Osaka, Japan \\
  \texttt{\{yamaguchi.takuya2015, ishii.yasunori, tsukizawa.sotaro\}@jp.panasonic.com} \\
}

\begin{document}

\maketitle

\begin{abstract}
Explaining predictions of deep neural networks (DNNs) is an important and nontrivial task. In this paper, we propose a practical approach to interpret decisions made by a DNN object detector that has fidelity comparable to state-of-the-art methods and sufficient computational efficiency to process large datasets. Our method relies on recent theory and approximates Shapley feature importance values. We qualitatively and quantitatively show that the proposed explanation method can be used to find image features which cause failures in DNN object detection. The developed software tool combined into the “Explain to Fix” (E2X) framework has a factor of 10 higher computational efficiency than prior methods and can be used for cluster processing using graphics processing units (GPUs). Lastly, we propose a potential extension of the E2X framework where the discovered missing features can be added into training dataset to overcome failures after model retraining.
\end{abstract}

\section{Introduction}
\label{sec:intro}
Recent achievements of deep neural networks (DNNs) make them an attractive choice in many computer vision applications including image classification~\citep{he} and object detection~\citep{speedacc}. At the same time, DNNs are often considered as \textit{black boxes} due to complex nonlinear interactions within their neural layers. This creates barriers for wide DNN adoption in critical applications. In practice, however, we are often interested in a more practical task of explaining DNN outputs to analyze and overcome failed predictions. To be specific, we consider a common object detection task and propose a method to analyze its outputs i.e. spatially estimate what caused a failure such that similar example can be added to a training dataset. Then, the retrained model with the augmented training dataset should not fail for such cases.

In this paper, we propose a method and software framework called “Explain to Fix” (E2X) to interpret decisions made by DNN object detector that has fidelity comparable to state-of-the-art methods and sufficient computational efficiency to process large datasets using graphics processing units (GPUs). Our method relies on recently proposed theory~\citep{shap} and approximates SHAP (Shapley additive explanations) feature importance values described in Section~\ref{sec:theory}. Next, we apply theoretical background to a popular single-shot object detector (SSD)~\citep{ssd} in Section~\ref{sec:ssd}. We qualitatively and quantitatively show advantages of the proposed explanation method in Section~\ref{sec:experiments}. Features discovered by the explanation model can be added into training dataset to fix failed cases in a manual or, potentially, in an automated fashion as proposed in Section~\ref{sec:future}. Unlike prior methods in Section~\ref{sec:related}, we employed the proposed explanation method to a more practically appealing object detection application that can be used in surveillance and autonomous driving systems.

\section{Related Work}
\label{sec:related}
Several types of methods to explain DNN predictions have been proposed recently. They can be classified by computation type into perturbation forward propagation approaches and backpropagation-based approaches. First research efforts were dedicated to prediction difference visualization methods where the part of input image is typically occluded (perturbed) by some patch and output change is visualized to present its importance. For example,~\citet{zeiler} used gray patches that usually match color distribution mean.~\citet{pda} extended this idea with the Probability Difference Analysis (PDA) method that samples patch color and location from Gaussian distribution for the corresponding color mean and center. The latter approach allows to analyze parts of the inputs which are close to distribution mean. The main drawback of perturbation methods is computational inefficiency since analysis is done for each input feature (e.g. pixel) by performing multiple per-feature inference passes.

Another direction was proposed by~\citet{simonyan} and further improved by~\citet{springenberg} where the input feature importance map was calculated by backpropagating gradients from the prediction into input space. Such backpropagation-based methods produce saliency maps that may explain DNN outputs. Compared to forward propagation methods, these methods have low computational complexity because a single backward pass obtains result for an entire input feature (pixel) space. At the same time, their main flaw was discovered by~\citet{montavon} who made a connection between explanation model and its first order Taylor series approximation. Other drawbacks of these methods are their unreliability due to nonlinear saturation, discontinuous and negative gradients. Next generation of backpropagation-based approaches addressed some of the drawbacks at the expense of higher computational complexity. The most notable methods are LRP~\citep{lrp}, DeepLIFT~\citep{deeplift}, and IG~\citep{ig}. 

One common drawback of the previously described approaches is their dependence on the underlying DNN model. It means that the \textit{explanation model} is not agnostic to the \textit{prediction model} which is untrustworthy. To address this drawback,~\citet{lime} proposed to explain any complex model by local interpretable explanation model using a method called LIME. In case of computer vision DNNs, the explanation model is an interpretable linear regression that operates over superpixels (patches) of images and tries to minimize $L_2$ difference between output of DNN model and linear explanation model. The drawbacks of this approach are reliance on superpixel segmentation algorithm, approximation using simplified linear model, and relatively high computational complexity.

One of the latest works by~\citet{shap} generalized most of the \textit{additive feature attribution methods} using a unified theoretical framework. Based on game theory, they proposed to estimate SHAP values as a measure of feature importance. This work showed that methods like LIME, LRP, and DeepLIFT can be viewed as approximate SHAP estimators. Due to reliance on game theory SHAP values guarantee certain important properties: local accuracy, missingness and consistency in explanations. Another recent analysis by~\citet{ancona} showed analytical connections between all backpropagation-based methods and their empirical similarities for MNIST image classification application both qualitatively and quantitatively. In particular, they concluded that backpropagation-based methods are nearly equivalent under some assumptions e.g. DNNs with ReLU nonlinearities. Unlike these prior works, E2X framework targets a more mature industry-level systems. Hence, we concentrate on further accuracy improvement, computational complexity optimization, practical object detection application and the potential use case scenario, where the explanation models can be applied to fix misses of DNN models.

\section{{E2X} Framework Development}
\label{sec:e2x}
\subsection{Theoretical Background}
\label{sec:theory}
Using the unified theory for \textit{local explanation methods} from~\citep{shap}, let $f(x)$ be the original \textit{prediction model} and $g(\acute{x})$ be the \textit{explanation model}, where $x$ is the input (image) and $\acute{x}$ is the simplified input (binary vector of features). The local methods expect $g(\acute{z}) \approx f(h_x(\acute{z}))$ whenever sample $\acute{z} \approx \acute{x}$. A mapping function $h_x(\acute{z}) = z$ represents superpixel segmentation for images. Using this notation, any \textit{additive feature attribution method} can be expressed as a linear function of binary variables
\begin{equation} \label{eq:1}
g(\acute{z}) = \phi_0 + \sum_{i=1}^{M} \phi_i \acute{z}_i,
\end{equation}
where $\acute{z}\in\{0,1\}^M$, $M$ is the number of simplified features and $\phi_i$ represents the $i$th feature importance value i.e. it attributes the output effect to the corresponding feature.

LIME segments input images $x$ into superpixels $h_x (\acute{x})$ and minimizes the $L_2$ loss
\begin{equation} \label{eq:2}
\sum_{z,\acute{z} \in Z} \pi_x(z) (f(z)-g(\acute{z}))^{2},
\end{equation}
where $\pi_x$ is exponential distance function. Eq.~(\ref{eq:2}) is solved using penalized linear regression to obtain $\phi_i$, where, typically, $M=O(2)$ or more and the binary vectors $\acute{z}$ are sampled $K$ times from uniform distribution. Hence, number of samples $K$ should be equal or more than $M$ to obtain a solution for~(\ref{eq:2}), which justifies the high computational complexity.

Backpropagation-based methods such as LRP, DeepLIFT and IG recursively attribute output effect to each input that changes from a reference $\tilde{x}$ to the original input $x$ as
\begin{equation} \label{eq:3}
\Delta f = f(x) - f(\tilde{x}) = \sum_{i=1}^M C_{\Delta x_i, \Delta f},
\end{equation}
where gradient-dependent $C_{\Delta x_i, \Delta f} = \phi_i \acute{x}_i$ and $f(\tilde{x}) = \phi_0$ from~(\ref{eq:1}). Therefore, such methods belong to additive feature attribution methods. The reference is usually a distribution mean i.e. grayscale background or it is selected heuristically. DeepLIFT, similarly to LRP, estimates gradient differences at each layer by progressively calculating weighted discrete gradient differences to the input image space. In contrast, the IG approach calculates gradient differences only at the input layer and approximates continuous gradients in $C_{\Delta x_i, \Delta f}$ by a discrete sum of gradients as a linear path from $\tilde{x}$ to $x$. Hence, both DeepLIFT and LRP require changing the processing of each layer, have issues with calculating discrete gradients for composite nonlinear functions, and produce different results for functionally the same networks. The latter method is more general but highly depends on approximation of gradient differences and reference-to-input mapping.

Recent work~\citep{shap} showed that explanation models from~(\ref{eq:1}) can approximate Shapley values i.e. feature importance values for any model in the presence of multicollinearity. They are a weighted average of all possible differences of combinations when a certain feature is present in the model or not. It is infeasible to calculate Shapley values and~\citet{shap} introduced SHAP values which assume feature independence, additivity and linearity of explanation model followed from~(\ref{eq:1}). In addition, this work generalized LIME and DeepLIFT by methods called Kernel SHAP and Deep SHAP, respectively. To efficiently approximate SHAP values, we select IG~\citep{ig} as a base approach to avoid the drawbacks of DeepLIFT and extend it with several new ideas. The base IG method estimates feature importance values as
\begin{equation} \label{eq:4}
\phi_i \approx \frac{(x_i - \tilde{x}_i)}{K} \sum_{k=1}^{K} \frac{\partial f (\tilde{x} + k(x - \tilde{x})/K)}{\partial (\tilde{x}_i + k(x_i - \tilde{x}_i)/K)},
\end{equation}
where $\tilde{x}$ is a reference image and $K$ is the number of steps to approximate continuous integral by a discrete sum. To make computation of~(\ref{eq:4}) tractable, usually all $M$ features are calculated simultaneously. Unfortunately, such simplification leads to noisy inaccurate results due to feature dependences.

We propose to extend original IG by introducing a mapping $u_x()$ similar to~\citep{lime} that, in addition, multiplies every segmented feature $h_x(\acute{x})_i$ by a weight $w_i$. The $h_x()$ mapping is justified by SHAP local feature independence assumption and converts noisy pixel-level estimates into more reliable superpixel-level values. The second drawback of the original IG is the input with the linear path between $x$ and $\tilde{x}$. It works when importance scores are estimated independently for each feature, but violates feature independence when all SHAP values are being simultaneously estimated. Then, even $h_x()$-segmented superpixels are not independent due to large DNN receptive fields. Instead, we propose to sample $M$ weight vectors $w_i(k)=k/(K-1)$, where $k\in\{0, 1, \ldots, K-1\}^K$ from uniform distribution. Hence, the linear path from~(\ref{eq:4}) is better approximated by sampled weights $w_i(k)$ for a segmented image. Assuming $\tilde{x}_i = E[x_i] = 0$ as in IG, SHAP values can be estimated by the proposed approach as 
\begin{equation} \label{eq:5}
\phi \approx s_x \left( \frac{x}{K} \sum_{k=1}^{K} \frac{\partial f(u_x(k))}{\partial u_x(k)} \right),
\end{equation}
where $K$ is the number of samples, $u_x(k)_i = w_i(k) h_x(\acute{x})_i$, and averaging mapping function $s_x(\acute{x})_i =  \sum_{h_i, w_i \in x_i} x_{h_i, w_i}/(H_i W_i)$. The latter mapping $s_x()$ averages estimated values for each $i$th segment. The proposed E2X method in~(\ref{eq:5}) extends IG method to estimate SHAP feature importance values assuming $h_x()$, which satisfies SHAP requirements, exists. 

\subsection{Extension to {SSD} Object Detector}
\label{sec:ssd}
In general, theoretical background from Section~\ref{sec:theory} can be applied to any DNN model. Unlike DeepLIFT or LRP, no change in the network architecture or processing is required for the E2X method in~(\ref{eq:5}). At the same time, often certain customization to a particular model is needed. In this paper, we evaluate the popular SSD object detector~\citep{ssd}. Unlike the simple image classification task employed in prior works with a single \textit{softmax} output, object detection calculates a collection of outputs that depend on spatial location and \textit{priorbox} properties. Moreover, SSD contains a number of layers where each activation represents a predefined bounding box of different aspect ratio and size. For example, the popular SSD300 model can output up to 8732 detections. To support this in the E2X framework, we modified the original SSD Caffe code to output the detection index within feature map activations in order to locate output features of interest.

The main objective of E2X is to identify and analyze failed object features. Failures in object detection are divided into false positives (FPs) and false negatives (FNs). When the groundtruth is available, FP indices can be easily identified because their bounding box coordinates do not sufficiently overlap with the groundtruth. Unfortunately, a more interesting FN class of detections is hard to identify in thousands of candidates because information about negatives \textit{a-priori} is missing in the output. We developed a special method to find potential FN candidates and analyze them. First, the output detection threshold is lowered. For example, the typical 50\% confidence threshold can be lowered to 1\%. Second, the Jaccard overlaps with groundtruth can be calculated for the subset of FN candidates. Lastly, the candidates with the maximum Jaccard overlap and confidence can be analyzed using E2X framework. Hence all FPs and most of the FNs can be analyzed using this procedure.

\section{Experiments}
\label{sec:experiments}
\subsection{Qualitative Results}
\label{sec:qual}
We apply E2X to the SSD300 model implemented in the Caffe framework~\citep{caffe} and pretrained on the VOC07+12 dataset~\citep{voc}. For comparison we selected the PDA, LIME and IG methods. All prior methods were reimplemented and are part of E2X framework which is publicly available\footnote{\href{https://github.com/gudovskiy/e2x}{https://github.com/gudovskiy/e2x}}. We do not compare with the older saliency map methods as well as recent LRP and DeepLIFT methods due to the reasons described in Sections~\ref{sec:related} and~\ref{sec:theory}. Lastly, we assume that backpropagation-based methods like IG and E2X calculate all feature importance values at once to reduce complexity. The selected methods were adopted with the same pre- and post-processing and executed on a single NVIDIA Quadro P6000 GPU.

\begin{figure}
	\centering
	\includegraphics[width=0.8\linewidth]{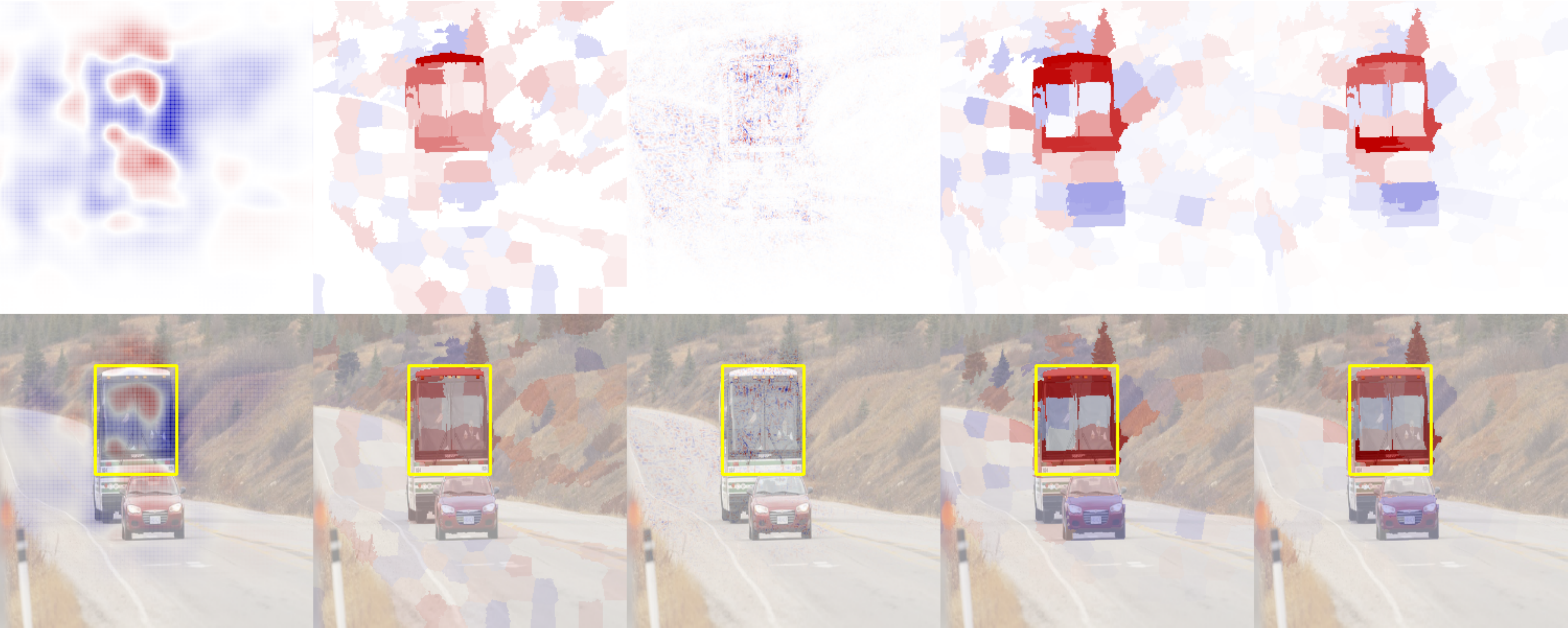}
	\caption{Qualitative comparison of the PDA, LIME($K=2000$), IG($K=32$), E2X($K=16$) and E2X($K=128$) explanation models depicted in the left to right order for bus false negative detection. Red and blue represent positive and negative features, respectively.}
	\label{fig:1}
\end{figure}

Figure~\ref{fig:1} illustrates a qualitative comparison of the feature importance maps at the top and the same maps overlaid over the input image at the bottom. Feature importance values are normalized to have unit (+/-1) range, where “+1” means the most positive feature (represented by red), “-1” means the most negative feature (represented by blue), and “0” means neutral feature (represented by transparent pixels). In this example, a bus object outlined by a yellow bounding box is being explained. This FN detection has only 48\% confidence due to several reasons which is revealed by being examined explanation models with the exception of IG. IG method does not clearly explain detected objects due to noisy per-pixel values. PDA highlights positively the most important bus areas but fails to identify negative class-discriminative features like the area occluded by a car. LIME and E2X clearly identify the car and some background as negative features. As $K$ grows, E2X produces less noisy estimates which follows from the decreased attention to the irrelevant surrounding areas. Hence, the E2X explanation model can robustly identify failing object features. Our evaluations show that FN detections are frequently caused by the nearby objects from another class or unseen background textures and FP detections are usually caused by random objects that look very similar to the training dataset examples.

\subsection{Quantitative Results and Computational Efficiency}
\label{sec:quant}
Quantitative comparison of explanation methods is nontrivial task since the explanation model with true Shapley values cannot be calculated. Hence, a certain explanation model could be selected as a reference. In our experiments, we select established LIME and PDA models with high-fidelity parameters for such references. The reference LIME model uses 1,000 "SLIC" segments and $K=10,000$. The reference PDA model is exact reimplementation of the original code but with the adaptive analysis window size (4$\times$ bounding box area) to make estimation time tractable. Furthermore, the PDA explanation models diverge from the reference model by adding striding (stride $S=\{4,8,16\}$) analysis with interpolation between missing points to decrease computations. The LIME explanation models use $K=\{200, 500, 2000\}$ samples and its segmentation method (“SLIC” with 200 segments) is the same as in the E2X models. The IG model always uses $K=32$ samples while the E2X models estimate importance maps with $K=\{8, 16, 32, 64, 128\}$.

\begin{figure}
	\centering
	\includegraphics[width=0.8\linewidth]{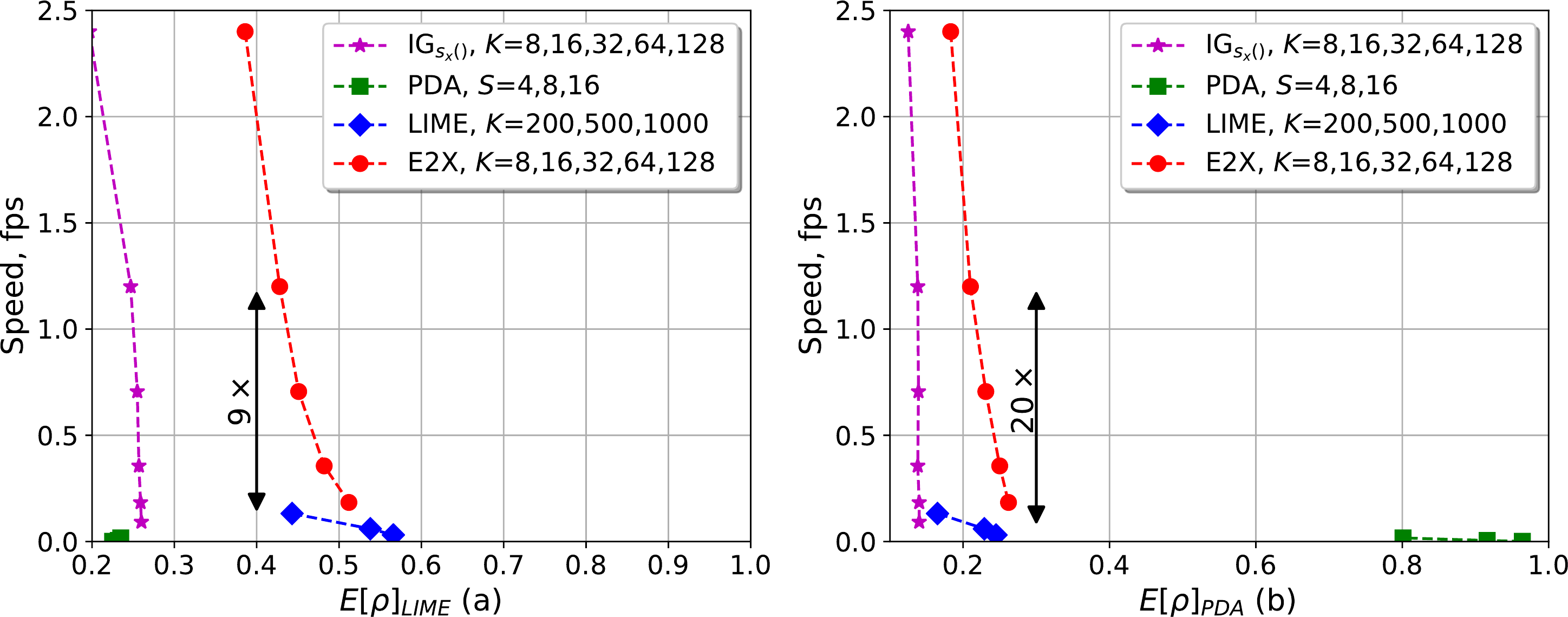}
	\caption{Correlation to reference model vs. computational speed.}
	\label{fig:2}
\end{figure}

For all experiments we selected 18 images with total 36 detections from the VOC07 test dataset with overlapping objects. Then, we calculate correlation coefficient $\rho$ between feature importance values of the reference model and each explanation model for every detection including true positives, FPs and FNs. Next, we average correlation coefficients and estimate $E[\rho]$. Figure~\ref{fig:2} illustrates $E[\rho]$ correlations vs. computational efficiency in terms of frames-per-second speed. Computational speed is estimated as the average wall-clock time needed to analyze a detection with a batch size of 32 for $K=32$ or 64 for $K>32$ using single GPU. 

Figure~\ref{fig:2} does not show results for IG due to very low $E[\rho]<0.01$ which confirms the qualitative result. But for an ablation study, we apply $s_x()$ from~(\ref{eq:5}) to IG. Such a modified IG$_{s_x()}$ emphasizes the importance of weight sampling in E2X. PDA results have relatively low correlation ($E[\rho]_{LIME}\approx0.2$) with the other methods. Most likely, this is due to spatially local \textit{probability difference} estimates that cannot capture global context. Model-agnostic LIME and the proposed E2X compute importance maps with relatively high correlation ($E[\rho]_{LIME}\approx0.5$). Although the LIME explanation models intrinsically have high bias towards LIME reference model, E2X generates very similar importance maps which can be explained by several reasons: both methods use the same segmentation method and, unlike PDA, their estimates are not spatially local. At the same time, E2X beats LIME in $E[\rho]_{PDA}$ which might be an indicator of a more precise explanation model. It was observed that LIME may put more attention to irrelevant areas due to nonorthogonal binary vectors in regression.

According to Figure~\ref{fig:2}, PDA's speed even with the proposed optimizations is very slow (less than 0.02 fps) compared with other methods. LIME's speed depends on the number of samples $K$ for GPU forward passes which are lower-bounded by the number of segments $M$ and CPU calculations to solve  $K \times M$ linear system. The speed of backpropagation-based E2X depends only on GPU capability and linearly scales with the selected $K$. Unlike LIME, E2X can be easily parallelized on multiple GPUs without having the CPU bottleneck. Using a single GPU, E2X is nearly 9$\times$ to 20$\times$ more computationally efficient than LIME with comparable feature importance map quality depending on the selected reference model.

\section{Use Case for Explanation Models}
\label{sec:future}
\begin{figure}
	\centering
	\includegraphics[width=0.4\linewidth]{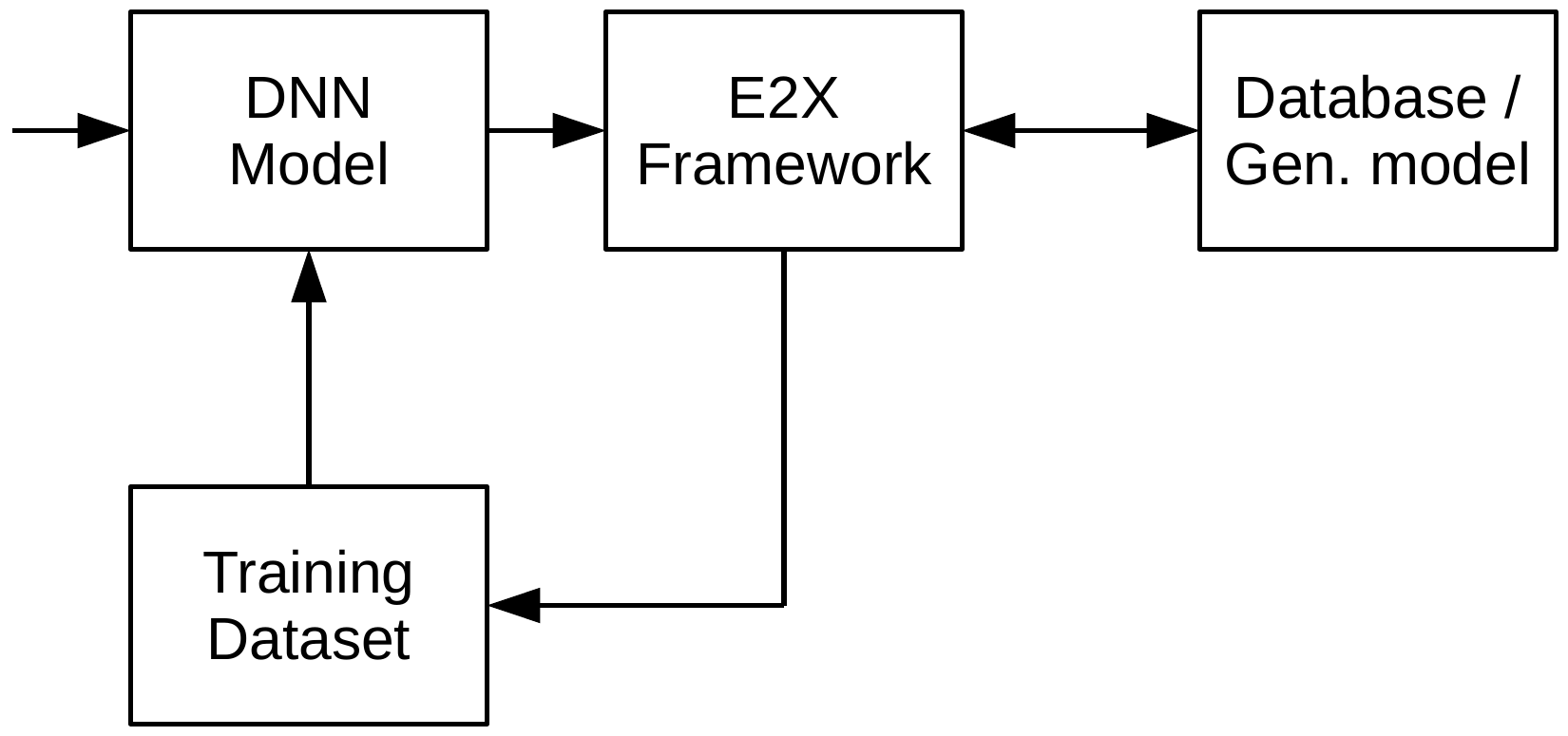}
	\caption{A concept of the automated dataset management solution using explanation model. E2X framework is extended by the method to query or synthesize missing samples in the training dataset.}
	\label{fig:3}
\end{figure}

Currently, explanation models are not widely utilized for commercial computer vision machine learning systems. At most they are used by an operator to verify decisions of a black box prediction model. This direction has limited usage and is not scalable due to cost factors. The developed E2X framework can reliably attribute DNN model predictions i.e. create an attention mask to failing or missing features in a reasonable amount of time. Then, such information can be used in an automated fashion to find which samples are missing in the training dataset. This can be done either by selecting them from a large database or synthesizing them using a generative model with the input conditioned on the identified features. An outline of such automated system is schematically shown in Figure~\ref{fig:3}. The concept is a closed-loop solution that allows to explain and fix not only current failures but also any future observed data.

\section{Conclusions}
\label{sec:conclusions}
In this paper, we reviewed recent methods to explain DNN model predictions and extended them for the task of object detection using the SSD detector. Next, we proposed a new method that combines high-fidelity explanations with the low computational complexity wrapped into the E2X software framework. This method approximately estimates recently proposed SHAP feature importance values using gradient backpropagation. We showed that the proposed explanation model qualitatively and quantitatively is comparable to state-of-the-art with a parametrized fidelity level. Experiments showed that computational complexity decreased by a factor of 10 for E2X compared with prior works. This makes it possible to process large datasets in reasonable time which is important for practical applications. Lastly, we identified a new use case scenario for explanation models and proposed a potential extension of the E2X framework to estimate failing or missing features in the training data which can be used as part of the automated dataset management system.

\bibliography{paper}

\end{document}